\documentclass{bmvc2k}
\usepackage{float}
\usepackage{graphics}
\usepackage{graphicx}
\usepackage{tabularx}
\usepackage{adjustbox}
\usepackage{nicematrix}
\usepackage{booktabs}
\usepackage{multirow}

\title{	
SAM-EG: Segment Anything Model with Egde Guidance framework for efficient Polyp Segmentation}

\addauthor{Quoc-Huy Trinh$^{1}$, Hai-Dang Nguyen $^{1}$, Bao-Tram Nguyen Ngoc}{}{1}
\addauthor{Debesh Jha$^{2}$, Ulas Bagci}{}{2}
\addauthor{Minh-Triet Tran}{}{1}

\addinstitution{University of Science, VNU-HCM\\ Ho Chi Minh, Vietnam}
\addinstitution{Northwestern University\\ Chicago, Illinois, USA}

\runninghead{Quoc-Huy Trinh, Minh-Triet Tran}{SAM-EG}

\begin{document}

\maketitle

\begin{abstract}

Polyp segmentation is a critical challenge in medical imaging, has prompted numerous proposed methods aimed at enhancing the quality of segmented masks. While current state-of-the-art techniques produce impressive results, the size and computational cost of these models pose challenges for practical industry applications. Recently, the Segment Anything Model (SAM) has been proposed as a robust foundation model, showing promise for adaptation to medical image segmentation. Inspired by this concept, we propose SAM-EG, a framework that guides small segmentation models for polyp segmentation to address the computation cost challenge. Additionally, in this study, we introduce the Edge Guiding module, which integrates edge information into image features to assist the segmentation model in addressing boundary issues from current segmentation model in this task. Through extensive experiments, our small models showcase their efficacy by achieving competitive results with state-of-the-art methods, offering a promising approach to developing compact models with high accuracy for polyp segmentation and in the broader field of medical imaging.

\end{abstract}

\section{Introduction}
\label{sec:intro}
Colorectal Cancer (CRC) stands as one of the most perilous diseases worldwide, emerging as a prevalent affliction affecting approximately one-third of the global population. To help the efficient treatment, the early diagnosis of CRC and Polyp segmentation tools are proposed to support the early treatment while this tool can help to localize the polyp region.

In recent years, numerous deep learning methods have been proposed to tackle this issue. UNet \cite{unet} initially introduced an encoder-decoder based architecture as an efficient approach for segmenting polyps in early stages. Subsequent methods, such as ResUNet \cite{resunet}, ResUNet++ \cite{jha2019resunet++}, UNet++ \cite{unet++}, DDANet \cite{dda} PEFNet \cite{pefnet}, and $M^{2}$UNet, propose improved versions of the encoder-decoder architecture to better capture the polyp object in endoscopic images. PraNet \cite{pranet}, one of the remarkable methods, employs supervision techniques to enhance the segmented polyp mask. Since then, several supervision-based methods \cite{pranet, polyp2seg, pvt} have been proposed, yielding positive results. State-of-the-art methods, including MetaPolyp \cite{metapolyp}, Polyp2SEG \cite{polyp2seg}, MEGANet \cite{bui2024meganet}, and Polyp-PVT \cite{pvt}, demonstrate impressive results and have the potential for further enhancement through real-world data application. However, these methods encounter challenges when deployed in real-world applications due to computational costs. For this reason, several compacted models such as ColonSegNet \cite{colonsegnet}, TransResUNet \cite{transresunet}, TransNetR \cite{TransNetR}, MMFIL-Net \cite{mmfilnet}, and KDAS \cite{kdas} have been proposed. Nonetheless, they also exhibit issues with boundary problems as mentioned in MEGANet \cite{bui2024meganet}, which affect the learning of the model on overall polyp shape, which can affect to the results when doing segmentation on small polyp objects.

Recently, the Segment Anything Model (SAM) \cite{sam} has emerged as a fundamental model for segmentation and has undergone extensive fine-tuning for medical segmentation tasks. Specifically, Med-SA \cite{Med-sA} and SAMed \cite{samed} have showcased promising results by training SAM's image encoder with a segmentation decoder. However, SAM's computational cost and its inability to provide boundary information are suboptimal for polyp segmentation. Despite this limitation, SAM can offer rich semantic knowledge for segmentation models, as highlighted by Julka et al. \cite{julka2023knowledge}.

To deal with previous challenges in Polyp Segmentation, we propose \textbf{SAM-EG}: \textbf{S}egment \textbf{A}nything \textbf{M}odel with \textbf{E}gde \textbf{G}uidance framework for efficient Polyp Segmentation, which is inspired by the SAM application. The primary goal of this framework is to guide the small segmentation model by using SAM semantic feature during the learning phase. By doing so, the small segmentation model can acquire semantic features from SAM, thereby enhancing its performance in the segmentation process. Moreover, to address boundary issues, we propose the Edge Guidance module (EG) to capture edge information from the input image and integrate it with the learned features from segmentation and SAM embedding. As a result, the segmentation model can prioritize edge information of the polyp, enriching boundary details during the model's learning process. In summary, our contributions are in three folds:

\begin{itemize}
    \item We propose SAM-EG, a guiding framework from the Segment Anything Model image encoder to the small model for efficient polyp segmentation.
    \item We introduce the Edge Guidance module (EG), to help the small model prioritize the edge information, which can leverage the boundary problem in polyp segmentation.
    \item We conduct extensive experiments to assess the performance and efficiency of our methods on diverse datasets, including Kvasir\cite{kvasir-seg}, Clinic-DB \cite{clinicdb}, Colon-DB\cite{colondb}, and Etis\cite{etis} dataset. 
\end{itemize}

This paper is organized as follows: in Section~\ref{sec:related}, we briefly review existing methods related to this research. Then we propose our methods in Section~\ref{sec:method}. Experiments setup are in Section~\ref{sec:exp}. Results of the experiment and the discussion are in Section~\ref{sec:result}. Finally, we present the conclusion in Section~\ref{sec:conclusion}.

\section{Related Work}
\label{sec:related}
\subsection{Polyp Segmentation}
Polyp segmentation is a task aimed at segmenting polyp objects from endoscopic images, thereby assisting doctors in making more accurate decisions during early diagnosis. The initial method addressing this challenge is UNet \cite{unet}, which employs an Encoder-Decoder architecture for segmentation. Subsequently, various methods based on the UNet architecture, such as UNet++\cite{unet++}, PEFNet \cite{pefnet}, and $M^{2}$UNet \cite{m2unet}, have emerged, focusing on improving feature extraction from the encoder and utilizing skip connections to address the limitations of boundary delineation in UNet. Additionally, PraNet \cite{pranet} introduces Supervision Learning as a novel approach to mitigate the sharpness boundary gap between a polyp and its surrounding mucosa. Building on this concept, subsequent methods, including MSNet \cite{MSNet}, have been proposed to further enhance the drawbacks associated with redundant feature generation at different levels of supervision. HarDNet-CPS \cite{hardnetcps} has been introduced to specifically concentrate on the lesion area. Recently, Polyp-PVT \cite{pvt} has been proposed to help suppress noises in the features and improve expressive capabilities, leading to significant results in polyp segmentation. While state-of-the-art methods such as MetaPolyp \cite{metapolyp}, Polyp2SEG \cite{polyp2seg}, MEGANet \cite{bui2024meganet}, and Polyp-PVT \cite{pvt} have demonstrated impressive results, the challenge remains in implementing these approaches in products due to the heavy computational demands of large models. In response to this issue, several methods have been proposed. For instance, ColonSegNet \cite{colonsegnet} introduces a lightweight architecture to mitigate the trade-off between prediction accuracy and model size. TransNetR \cite{TransNetR} are Transformer-based architectures that incorporate outlier detection methods, enhancing the generalization of lightweight models. Additionally, MMFIL-Net \cite{mmfilnet} proposes a lightweight model with integrated modules to address issues related to varying feature sizes in lightweight models, and KDAS \cite{kdas} proposes a Knowledge Distillation framework with Attention mechanism for efficient Polyp Segmentation. While these methods show promising initial results, they still exhibit a significant performance gap compared to state-of-the-art models and the lack of learning of the boundary information, which can affect the segmentation result of the model (pointed out by MEGANet \cite{bui2024meganet}  and KDAS \cite{kdas}). 

To alleviate previous works limitations, we propose \textbf{SAM-EG}, a framework that employs the Segment Anything Model to guide the segmentation model during the learning stage. Moreover, we incorporate Edge information via the edge detector, especially by using the Sobel edge detector to help the model prioritize the edge of the polyp, thus aiding it in learning the boundary information of the polyp tissue.

\subsection{Segment Anything Model (SAM)}

The Segment Anything model has become a popular foundation for prompting image segmentation. However, the robustness and generality of the image encoder from SAM enable its use in a variety of segmentation tasks, particularly in medical image segmentation. Several methods, such as SAM3D \cite{SAM3D}, SAM-UNETR\cite{SAM-UNETR}, AFTer-SAM \cite{AFTerSAM}, Med-SA \cite{Med-sA}, and SAMed\cite{samed} propose methods that employ the image encoder of SAM by fine-tuning it with a decoder for segmentation. These methods have shown promising results in medical segmentation and can be further improved for real-world scenarios. However, the main challenge of using the full image encoder from SAM lies in the computational cost during the inference stage, which poses difficulties with hospital hardware requirements.

In addition, our exploration reveals that the semantic knowledge from the SAM image encoder can be transferred to the segmentation model to enhance its segmentation learning by guiding the semantic features to the segmentation model. Inspired by this discovery, SAM-EG employs the SAM image encoder to guide the segmentation model in the framework for polyp segmentation. Furthermore, it is notable that the basic features from the Segment Anything Model lack boundary features, which may hinder the segmentation model's ability to segment polyps in endoscopic images. This is why we introduce the EG module, which incorporates edge features to enhance boundary information in the image features from SAM when guiding the segmentation model. This enables our framework to effectively address this challenge.
\section{Method}
\label{sec:method}
\begin{figure}[ht]
    \centering
    \includegraphics[width=0.8\linewidth]{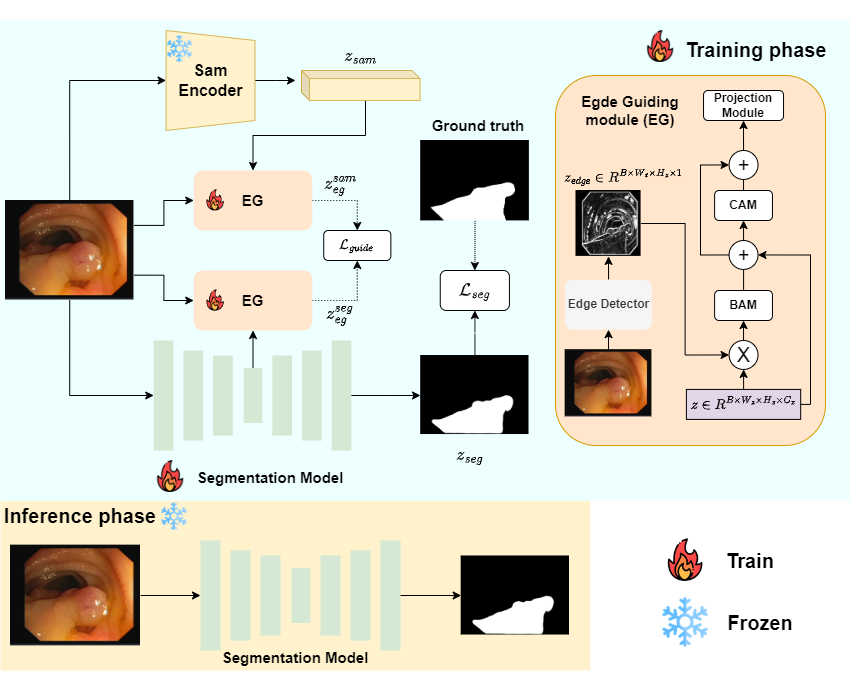}
    \caption{General SAM-EG framework}
    \label{fig:sameg}
    \vspace{-5mm}
\end{figure}

In this section, we introduce the overall SAM-EG framework, as illustrated in Figure~\ref{fig:sameg}. The framework consists of three main parts: the SAM image Encoder model (Sam Encoder), the Segmentation model, and the Edge Guiding Module. Within this framework, the SAM Encoder serves as a teacher model, transferring knowledge to the segmentation model, which acts as the student model. Additionally, we introduce the Edge Guiding module (EG) to address the boundary problem mentioned in MEGANet \cite{bui2024meganet}.

The input images $X$ with shape $B \times 352 \times 352 \times 3$ (with $B$ is the batches) are processed through the SAM Encoder $f_{sam}$ and the Segmentation Model $f_{seg}$. Notably, during the training phase, $f_{sam}$ remains frozen, while the $f_{seg}$ is unfrozen. Subsequently, the global features from the SAM Encoder $z_{sam}$ and the Segmentation model $z_{seg}$ undergo guidance through our Edge Guiding Module $f_{eg}$ to form two global features $z^{sam}_{eg}$ and $z^{seg}_{eg}$ before the knowledge transfer process begins. The $f_{eg}$ modules are learned during the training phase.

\vspace{-2mm}
\vspace{-3mm}
\subsection{SAM Encoder}

In the SAM Encoder branch, we utilize the image encoder from the Segment Anything Model (SAM), which is a pretrained using the MAE Vision Transformer \cite{vitmae}. This image encoder takes the input image $X$ and generates global features  $z_{sam} \in R^{B \times 16 \times 16 \times 256}$. Leveraging this encoder allows us to benefit from a well-trained model on a large dataset, thereby producing rich-information semantic features for guiding the segmentation model.
\vspace{-2mm}
\subsection{Segmentation Model}

In the Segmentation Model ($f_{\text{seg}}$ branch), we adopt the Polyp-PVT based architecture \cite{pvt} with the PVTV2-B0 backbone, which is the smallest version. Initially, the backbone encoder encodes the image $X$ into four scale features $f_{i}^{h} \in R^{\frac{W}{2^{i+1}} \times \frac{H}{2^{i+1}} \times C_{i}}$, where $C_{i} \in \{64, 128, 320, 512\}$, and $i \in \{1,2,3,4\}$, with $W$ and $H$ being the dimensions of the input image. Afterward, the features in the last scale of the encoder are utilized for guiding with the SAM encoder, while the remaining features are forwarded to the proposed modules in the Polyp-PVT \cite{pvt} architecture to obtain segmentation results. In addition, in the inference stage, just only small segmentation model is used to segment the polyp object.

\subsection{Edge Guiding module (EG)}
As point out in MEGANet \cite{bui2024meganet}, the main challenge of the polyp segmentation task is the boundary problem. Additionally, the guiding through SAM just help the segmentation model learn the basis feature, which is the pattern from polyp, not the general polyp shape. This can make the segmentation model lack focus on the boundary information, leads to the bad result in the segmentation. To deal with this challenge the Edge Guiding module (EG) $f_{eg}$ is to help the segmentation learn and prioritize the boundary information of the image.

 \noindent This module takes two inputs: the input images and the input feature $z \in R^{B  \times W_{z} \times H_{z} \times C_{z}}$, where $W_{z}$, $H_{z}$, and $C_{z}$ represent the width, height, and channel of the input feature respectively. The input images are initially converted to grayscale images and passed through the Edge Detector to generate the edge features, and these edge features are resized to $z_{edge} \in R^{B \times W_{z} \times H_{z} \times 1}$ (the effect of the different Edge Detectors is depicted in the Section~\ref{sec:ablation}). To fuse the edge feature with the image feature, we first resize the edge feature to match the shape of the image feature $z$, and then perform element-wise multiplication between the edge feature and the image feature $z$ to produce the fused feature $z_{fused}$. Subsequently, the Boundary Attention Module (BAM) and Channel Attention Module (CAM) are applied to form the $z_{bam}$ and $z_{cam}$ features, which focus on vital parts of the image feature and edge feature, thus improve the prioritize the boundary information, particularly the polyp region. Moreover, between each module, the skip connection is applied to mitigate the vanishing gradient. Finally, the Projection Module is utilized to transform these features into embeddings for the guiding process. 

\noindent The three modules, BAM, CAM, and Projection Module, are described as follows.

\noindent\textbf{Boundary Attention Module (BAM):} From our experiments, we observed that the edge feature may contain noise, which can adversely affect the learning process. This is the motivation behind our introducing the Boundary Attention Module (BAM), as depicted in Equation~\ref{equa:bam}.

\begin{equation}
    z_{bam} = z_{fused} \otimes \sigma(Conv(z_{fused}))  
    \label{equa:bam} 
\end{equation}

The fused edge feature \( z_{fused} \) is initially extracted through a convolution operation with sigmoid activation \( \sigma(.) \), forming the attention mask. Subsequently, the multiplication operation between \( z_{fused} \) and the attention mask is applied to create \( z_{bam} \). This approach enables our model to focus on the vital region, which is the polyp, instead of learning from the noisy background.

\noindent\textbf{Channel Attention Module (CAM):} 
To highlight the edges and important features from the fused results, we incorporate the Channel Attention Module (CAM). Within this block, the input from the previous stage, \( z_{bam} \), aggregates spatial information from a feature map through both average-pooling and max-pooling operations, generating two distinct spatial context descriptors. These descriptors are subsequently fed into a shared network to produce our channel attention map, which highlights the importance. The resulting channel attention maps are then multiplied with \( z_{bam} \) to generate the output feature, \( z_{cam} \), that focus on the important features of the polyp, as illustrated in Equation~\ref{equa:cam}.

 \begin{equation}
     z_{cam} = z_{bam} \otimes \sigma(W_{1}(W_{0}(AvgPool(z_{bam}))) + W_{1}(W_{0}(MaxPool(z_{bam}))))
     \label{equa:cam}
 \end{equation}

Where \(\sigma\) denotes the sigmoid activation function. Following the first fully connected layer, the ReLU activation is also applied. 

By applying this module, the model can prioritize the crucial parts from the fusion of image feature map and the edge information, which can benefit it in learning the polyp general feature and the boundary information from the input image.

\noindent \textbf{Projection Module:} The primary objective of the Projection Module is to convert the final feature map, after passing through CAM, into an embedding for guiding between SAM and the Segmentation model. Given the input $z_{\text{cam}}$, the output of this module $z_{\text{eg}}$ represents the fused feature with edge information. In this stage, $z_{\text{cam}}$ undergoes several operations, including global average pooling, linear projection, batch normalization, and the non-linear ReLU activation function. These operations generate an implicit embedding representing the polyp features with shape $B \times d$ (The $d$ value denotes the embedding dimension, and it is 256 in our implementation).

\subsection{Objective function}
The overall objective function for this framework (depicted in Equation~\ref{equa:lkdas}) is the sum of $\mathcal{L}_{guide}$ from the knowledge transfer of SAM embedding ($z_{eg}^{sam}$) with segmentation embedding ($z_{eg}^{seg}$), and the $\mathcal{L}_{seg}$ between the segmentation prediction $z_{seg}$ and the ground truth $y$. For the guide loss, we employ the L2 Loss, as demonstrated in Equation~\ref{equa:lguide}. Meanwhile, the segmentation loss ($\mathcal{L}_{seg}$) is the combination of binary cross-entropy loss ($\mathcal{L}_{BCE}$) and the dice loss ($\mathcal{L}_{dice}$), as depicted in Equation~\ref{equa:lseg}. 

\begin{gather}
    \mathcal{L}_{guide} = \frac{1}{N}||z_{eg}^{seg} - z_{eg}^{sam}||^{2}_{2}
    \label{equa:lguide}
    \\
    \mathcal{L}_{seg} =  \sum^{D}_{i}\mathcal{L}_{BCE}(z_{seg}^{i}, y^{i}) +  \mathcal{L}_{dice}(z_{seg}^{i}, y^{i}) 
    \label{equa:lseg}
    \\
    \mathcal{L}_{total} = \mathcal{L}_{guide}  +  \mathcal{L}_{seg}
    \label{equa:lkdas}
\end{gather}

\noindent Which $D$ denotes the number of decoded layers, and $N$ is the number of samples. In the implementation of Polyp-PVT \cite{pvt}, the $D$ value is equal to 2. 

By minimizing the guide loss, the segmentation model can learn the semantic feature from the SAM model. Additionally, the integration of the Edge Guiding module can help the transfer information contain and prioritize the boundary information, which impact beneficially to the segmentation model. 

\section{Experiment}
\label{sec:exp}
\subsection{Datasets}
To conduct the fair comparison, the experiment's dataset follows the merged dataset from the PraNet \cite{pranet} experiment for the training stage which includes 900 samples from Kvasir-SEG \cite{kvasir-seg} and 550 samples from CVC-ClinicDB \cite{clinicdb}. The remaining images of Kvasir-SEG \cite{kvasir-seg} and CVC-ClinicDB \cite{clinicdb} with two unseen datasets such as ColonDB \cite{colondb}, and ETIS \cite{etis} are used for benchmarking our method.

\subsection{Implementation Detail}

In our implementation, we conducted experiments based on the Polyp-PVT baseline \cite{pvt}. We utilized the PyTorch framework and employed a Tesla H100 80GB for training. The images were resized to $352 \times 352$, and a batch size of 16 was set. The AdamW optimizer was employed with a learning rate of $1e-4$, and weight decay value is $1e-4$. The best weights were obtained after 100 epochs, with a total training time of approximately 2 hours. During testing, the images were resized to $352 \times 352$.  

\subsection{Evaluation Metrics}

Two commonly used evaluation metrics, mean Dice (mDice) and mean Intersection over Union (mIoU), are employed for assessment. Higher values for both mDice and mIoU indicate better performance. In the context of polyp segmentation, the mDice metric holds particular significance as it is considered the most important metric for determining the effectiveness of a model. Moreover, we also take the comparison in the number of parameters, and FLOPs values, to illustrate the efficiency and light-weight of our method.

\subsection{Performance Comparisons}
In order to assess the effectiveness of our method, we compare our method with the state-of-the-art methods, and the real-time approaches.

\textbf{Comparison with State-of-the-art} To assess the effectiveness of our model, we compare our PVTV0 distilled model (approximately 3.7 million parameters) with several methods that have a higher parameter count. These include UNet \cite{unet}, UNet++ \cite{unet++},  PraNet \cite{pranet}, MSNet \cite{MSNet}, PEFNet \cite{pefnet}, $M^2$UNet \cite{m2unet}, and HarDNet-CPS \cite{hardnetcps}. In this comparison, we also provide information about the number of parameters to evaluate the impact of model size on overall performance. As the datasets used in PEFNet \cite{pefnet} differ, we retrain this method in our dataset setting, which follows PraNet\cite{pranet} for a fair and consistent comparison.

\textbf{Comparison with real-time methods:}  To evaluate the performance of our small model in terms of both prediction accuracy and computational efficiency, we conducted comparisons with several methods, namely ColonSegNet\cite{colonsegnet},  TransNetR\cite{TransNetR}, MMFIL-Net\cite{mmfilnet}, and KDAS \cite{kdas}. Except for MMFIL-Net, we reproduced the training processes for the remaining models using the same training dataset and testing dataset as our models. The comparison weights were carefully selected to ensure a fair and meaningful comparison.
\vspace{-5mm}
\section{Result}
\label{sec:result}
\subsection{Qualiative Comparison}
\noindent\textbf{Comparison with State-of-the-art:} As shown in Table~\ref{table:quality}, SAM-EG enhances the performance of the small model across all benchmark datasets, despite having the lowest number of parameters. Particularly, in the ColonDB dataset, our method surpasses by $+1.9$ and $+1.1$ in mDice and mIoU respectively. In the ETIS dataset, our method outperforms by $+3.8$ and $+1.7$ in mDice and mIoU respectively. These findings highlight the generalization ability of our small model in both familiar and unfamiliar domains, despite its substantially lower number of parameters compared to existing methods. Additionally, these results highlight the ability of the model in localize the small polyp objects, which is the limitation of previous works. This observation underscores the promising results of our contribution in creating a small model, demonstrating its accuracy in predictions comparable to that of larger models.

\begin{table*}[htbp]
\tiny
\centering

\begin{tabular}{@{}lcccccccccc@{}}
\toprule
\multirow{2}{*}{Method} & \multirow{2}{*}{Params(M)} & \multicolumn{2}{c}{ClinicDB}    & \multicolumn{2}{c}{ColonDB}     & \multicolumn{2}{c}{Kvasir}      & \multicolumn{2}{c}{ETIS}               \\
                              &      & mDice          & mIoU           & mDice          & mIoU           & mDice          & mIoU           & mDice     & mIoU            \\ \midrule
UNet (2015)  \cite{unet}     &   7.6     & 0.824 & 0.767      & 0.519     & 0.449                           & 0.821        & 0.756  & 0.406         & 0.343                    \\
UNet++ (2018)   \cite{unet++} &  9.0   & 0.794  & 0.729         & 0.483          & 0.410             & 0.820       & 0.743             & 0.401      & 0.344                          \\
PraNet (2019) \cite{pranet} & 32.6 & 0.899 & 0.849 & 0.712  & 0.640 & 0.898   & 0.840 & 0.628  & 0.567 
\\
MSNet (2021) \cite{MSNet} & 29.7 & \underline{0.921} & \underline{0.879} & \underline{0.755}  & 0.678 & 0.907 & \textbf{0.862} & \underline{0.719} & \underline{0.664}  
\\
PEFNet (2023) \cite{pefnet} & 28.0 & 0.866 & 0.814 & 0.710  & 0.638 & 0.892 & 0.833 & 0.636 & 0.572 
\\
$M^2$UNet (2023) \cite{m2unet} & 28.7 & 0.901 & 0.853 & 0.767  & \underline{0.684} & 0.907 & 0.855 & 0.670 & 0.595\\
HarDNet-CPS (2023) \cite{hardnetcps} & $--$ & 0.917 &  \textbf{0.887}  & 0.729 & 0.658 & \underline{0.911}  & \underline{0.856} &  0.69 & 0.619 
\\
\hline
\textbf{SAM-EG} & \textbf{3.7} & \textbf{0.931} & \underline{0.879} & \textbf{0.774} & \textbf{0.689} & \textbf{0.915}  & \textbf{0.862} & \textbf{0.757} & \textbf{0.681}  &  \\ \bottomrule
\end{tabular}
\caption{Qualiative results of \textbf{SAM-EG} on various datasets}
\label{table:quality}
\end{table*}

\noindent\textbf{Comparison with real-time methods:} As depicted in 
In Table~\ref{table:realtime}, SAM-EG outperforms previous real-time models across four different datasets and in all metrics. Despite having similar numbers of parameters and FLOPs value as KDAS, our method achieves gradually better performance on all four datasets. Particularly notable in the ColonDB dataset, our method surpasses the second best method by $+1.5$ and $+1.0$ in mDice and mIoU metrics respectively. These results indicate that the strength of the SAM image encoder can benefit the learning of the segmentation model, enabling it to address the challenges of polyp segmentation and achieve competitive results.
\begin{table*}[htbp]
\tiny
\centering

\begin{tabular}{@{}lcccccccccccc@{}}
\toprule
\multirow{2}{*}{Method} & \multirow{2}{*}{Params(M)}  & \multirow{2}{*}{FLOPs (G)}  & \multicolumn{2}{c}{ClinicDB}    & \multicolumn{2}{c}{ColonDB}     & \multicolumn{2}{c}{Kvasir}      & \multicolumn{2}{c}{ETIS}               \\
                               & &            & mDice          & mIoU           & mDice          & mIoU           & mDice          & mIoU           & mDice      & mIoU         \\ \midrule
ColonSegNEt (2021) \cite{colonsegnet} & 5.0 & 6.22 & 0.28 & 0.21 & 0.12 & 0.10 & 0.52 & 0.39  & 0.16 & 0.12\\
TransNetR (2023) \cite{TransNetR} & 27.3 & 10.09   & 0.87 & 0.82 & 0.68 & 0.61  & 0.87 & 0.80 & 0.60 & 0.53  \\
MMFIL-Net (2023) \cite{mmfilnet} & 6.7 & 4.32 & 0.890 & 0.838  & 0.744 & 0.659 & 0.909 & \underline{0.858} & 0.743  & 0.670 \\
KDAS (2024) \cite{kdas} & 3.7 & 2.01 & \underline{0.925} & \underline{0.872} & \underline{0.759} & \underline{0.679} & \underline{0.913}  & 0.848 & \underline{0.755} & \underline{0.677}  &  \\ 
\hline
\textbf{SAM-EG} & \textbf{3.7} & \textbf{2.12} & \textbf{0.931} & \textbf{0.879} & \textbf{0.774} & \textbf{0.689} & \textbf{0.915}  & \textbf{0.862} & \textbf{0.757} & \textbf{0.681}  &  \\ \bottomrule
\end{tabular}
\caption{Comparison of  model from \textbf{SAM-EG} with real-time model}
\label{table:realtime}
\end{table*}
\vspace{-3mm}
\subsection{Qualitative visualization}

\begin{figure}[ht]
    \centering
    \includegraphics[width=0.8\linewidth]{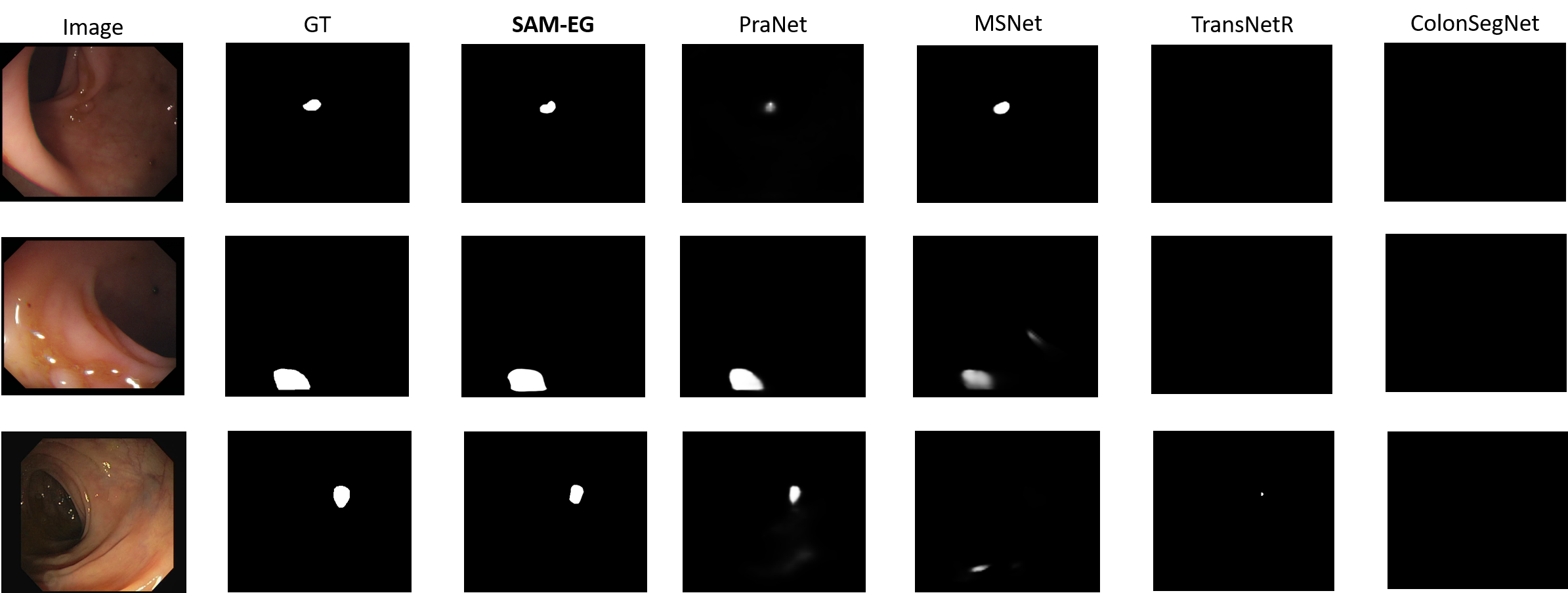}
    \vspace{-2mm}
    \caption{Visualization comparison between SAM-EG with several methods}
    \label{fig:vis}
    \vspace{-3mm}
\end{figure}

Figure~\ref{fig:vis} illustrates a comparison of segmentation between SAM-EG and other methods, both state-of-the-art and real-time. The visualization demonstrates that our method effectively identifies difficult and tiny polyps from endoscopic images. This indicates that by integrating guidance from SAM and learning from edge information, our model can capture boundary information, thus achieving promising results.

\subsection{Ablation Studies}
\label{sec:ablation}
\textbf{The effect of SAM Guiding and EG Module:} To assess the effectiveness of the SAM Guiding and EG Module, we conducted experiments in three scenarios. The first scenario involved training Polyp-PVT with backbone PVTV2-B0 without using SAM for guiding or the Edge Guiding Module. In the second scenario, SAM was used for guiding but the EG module was not included. The final scenario represents our full SAM-EG approach. Results are summarized in Table~\ref{table:sameg}. From the results, we observe a gradual improvement by incorporating SAM Guiding and the Edge Guiding module, highlighting the importance of edge information and the benefits of guiding segmentation with semantic features from the foundational model.
\vspace{-2mm}
\begin{table*}[htbp]
\small
\centering

\begin{tabular}{@{}lcccccccccccc@{}}
\toprule
\multirow{2}{*}{Method}  & \multicolumn{2}{c}{ColonDB}     & \multicolumn{2}{c}{Kvasir}    \\
                              & mDice          & mIoU           & mDice          & mIoU               \\ \midrule
\textbf{Baseline}  & 0.756 & 0.668 & 0.894  & 0.835 \\
\textbf{Sam Guiding + w/o EG} & 0.766 & 0.673 & 0.899 & 0.841 \\
\textbf{SAM-EG}  & \textbf{0.774} & \textbf{0.689} & \textbf{0.915}  & \textbf{0.862} \\ \bottomrule
\end{tabular}
\caption{Effect of SAM guiding and the EG module}
\label{table:sameg}
\end{table*}

\noindent\textbf{Effect of Edge Detector:} To assess the effectiveness of the Edge Detector within the entire pipeline, we conducted experiments to compare results using three different detectors: Canny, Laplacian, and Sobel. The outcomes are summarized in Table~\ref{table:edetector}, where the Sobel detector yielded the best performance. This suggests that the Sobel Edge detector is well-suited for the task of polyp segmentation. The reason behind this result is due to the lower sensitivity to noise of the Sobel operator, which allows its edge information to emphasize the edges of polyp tissue more effectively.
\begin{table*}[htbp]
\small
\centering

\begin{tabular}{@{}lcccccccccccc@{}}
\toprule
\multirow{2}{*}{Method}  & \multicolumn{2}{c}{ColonDB}     & \multicolumn{2}{c}{Kvasir}    \\
                              & mDice          & mIoU           & mDice          & mIoU               \\ \midrule
\textbf{Canny detector}  & 0.772 & 0.687 & 0.898  & 0.842 \\
\textbf{Laplacian detector} & 0.772  & 0.689 & 0.901 &  0.853  \\
\textbf{Sobel detector}  & \textbf{0.774} & \textbf{0.689} & \textbf{0.915}  & \textbf{0.862} \\ \bottomrule
\end{tabular}
\caption{Comparison the effect of the Edge Detector to the segmentation results}
\label{table:edetector}
\end{table*}
\vspace{-5mm}
\section{Conclusion}
\vspace{-3mm}
\label{sec:conclusion}
In conclusion, we introduce SAM-EG, a framework that guides a small model using the Segment Anything Model (SAM) along with edge information to address the computational costs and boundary challenges in polyp segmentation for real-world applications. By the benefit of guiding semantic features from SAM, the segmentation model can improve the segmentation result while incorporating edge information, our segmentation model learns boundary details, enabling it to address challenges posed by difficult polyps such as tiny tissues or hard-to-identify areas. Through extensive experiments, our method achieves competitive results compared to state-of-the-art and real-time models, despite employing a small number of parameters and FLOPs. These results demonstrate the potential of our approach for real-world applications. 

In the future, we aim to explore more efficient mechanisms and encourage researchers to delve into this topic, which could greatly benefit the integration of polyp segmentation applications into clinical environments.

\bibliography{egbib}
\end{document}